\pdfoutput=1
\documentclass[11pt]{article}

\usepackage[]{emnlp2021}

\usepackage{times}
\usepackage{latexsym}
\usepackage{amsmath}
\usepackage{graphicx}
\newcommand\blfootnote[1]{%
  \begingroup
  \renewcommand\thefootnote{}\footnote{#1}%
  \addtocounter{footnote}{-1}%
  \endgroup
}

\usepackage[T1]{fontenc}
\usepackage[utf8]{inputenc}

\usepackage{microtype}
\usepackage{todonotes} % for comments

\usepackage{xcolor}

\title{Semi-Supervised Few-Shot Intent Classification and Slot Filling}

\author{Samyadeep Basu*,  Karine lp Kiun Chong*, Amr Sharaf*,  Alex Fischer, Vishal Rohra, \\
\bf{Michael Amoake, Hazem El-Hammamy, Ehi Nosakhare, Vijay Ramani, Benjamin Han} \\
\{sbasu, kaipkiun, amrsharaf, alex.fischer, virohra, miamoako, haelhamm, \\
ehnosakh, vijayram, diha\}@microsoft.com
\\
\bf{Microsoft AI}}
\usepackage[normalem]{ulem} %to use strikethrough (sout)

\begin{document}
\maketitle

\begin{abstract}

Intent classification (IC) and slot filling (SF) are two fundamental tasks in modern Natural Language Understanding
(NLU) systems. Collecting and annotating large amounts of data to train deep learning models for such systems is not
scalable. This problem can be addressed by learning from few examples using fast supervised meta-learning techniques
such as prototypical networks. In this work, we systematically investigate how contrastive learning and unsupervised
data augmentation methods can benefit these existing supervised meta-learning pipelines for jointly modelled IC/SF
tasks. Through extensive experiments across standard IC/SF benchmarks (SNIPS and ATIS), we show that our proposed
semi-supervised approaches outperform standard supervised meta-learning methods: contrastive losses in conjunction 
with prototypical networks consistently outperform the existing state-of-the-art for both IC and SF tasks, while data
augmentation strategies primarily improve few-shot IC by a significant margin.

% TODO Add numbers / results from experiments

\end{abstract}

\section{Introduction}
We study the problem of few-shot Intent Classification (IC) and Slot Filling (SF). In the few-shot learning setting, the
learner has to learn given only a handful of training examples. We propose a semi-supervised approach for solving this
problem based on augmenting supervised meta-learning with unsupervised data augmentation and contrastive learning. We
systematically investigate how different data augmentation and contrastive learning strategies improve IC/SF
performance, and show that our semi-supervised approach outperforms state-of-the-art models for few-shot IC/SF.
\blfootnote{* First three authors contributed equally}
Given the user utterance: ``\emph{Book me a table for 6 at Lebanese Taverna}'', an IC model identifies
``\emph{Restaurant Booking}'' as the intent of interest, and an SF model identifies the slot types and values:
\emph{Party\_Size:"6", Name: "Lebanese Taverna"}. These functionalities are typically driven by powerful deep learning
models that rely on huge amounts of domain specific training data. As such labeled data is rarely available, building
models that can learn from only a few examples per class is inevitable.

Few-shot learning techniques~\citep{krone2020learning, few_triplet} have been recently proposed to address the problem
of generalizing to unseen classes in IC/SF when only a few training examples per class are available.
\citet{krone2020learning} utilized meta-learning approaches such as prototypical networks \citep{Snell} and MAML
\citep{pmlr-v70-finn17a} to jointly model IC/SF. They showed that prototypical networks outperform other prevalent
meta-learning techniques such as MAML as well as fine-tuning. In this paper, we extend this powerful supervised
meta-learning technique with unsupervised contrastive learning and data augmentation.

\citet{rajendran2020metalearning} showed that meta-learners can be particular prone to overfitting which can be
partially alleviated by data augmentation \citep{liu2020task}. Data augmentation strategies in NLP have been shown to
boost performance in general text classification settings \citep{DBLP:journals/corr/abs-1901-11196,
DBLP:journals/corr/abs-1904-12848, lee2021neural}, however, there exists very little work on how data augmentation can
be effectively used in the meta-learning pipeline specific to NLU tasks. To address this question, we first introduce a
novel data augmentation strategy \texttt{slot-list values} for IC/SF tasks which generates synthetic utterances using
dictionary-based slot-values. Additionally, we investigate how state-of-the-art augmentation strategies such as
backtranslation \citep{DBLP:journals/corr/abs-1904-12848} and perturbation-based augmentations such as EDA --
Easy Data Augmentation \cite{DBLP:journals/corr/abs-1901-11196} -- can be used alongside prototypical networks.

We further investigate how contrastive learning \citep{chen2020simple} can be used as an additional regularizer during
the meta-training stage to create better generalizable meta-learners. Contrastive learning is useful in creating
improved prototypes as they pull similar representations together while pushing apart dissimilar ones. Through extensive
experiments across SNIPS and ATIS, we show that meta-training with contrastive losses in conjunction with the general
prototypical loss function improves IC/SF performance for unseen classes with few examples. Our contributions include: 

\begin{itemize}
   \item We demonstrate the effectiveness of contrastive losses as a regularizer in the meta-learning pipeline, by
       empirically showing how it improves few-shot IC/SF tasks across benchmark datasets such as SNIPS and ATIS.
    \item We illustrate the positive impact of data augmentation techniques such as backtranslation and EDA in improving few-shot IC tasks.
\end{itemize}

\section{Proposed Approaches}

\begin{table*}[t]
\hspace{-1em}
\scalebox{0.58}{
\begin{tabular}{|c|c|c|c|c|c|l|l|l|l|}
\hline
 & Level & \multicolumn{2}{c|}{\begin{tabular}[c]{@{}c@{}}SNIPS\\ (Kmax=20)\end{tabular}} & \multicolumn{2}{c|}{\begin{tabular}[c]{@{}c@{}}ATIS\\ (Kmax=20)\end{tabular}} & \multicolumn{2}{c|}{\begin{tabular}[c]{@{}c@{}}SNIPS\\ (Kmax=100)\end{tabular}} & \multicolumn{2}{c|}{\begin{tabular}[c]{@{}c@{}}ATIS\\ (Kmax=100)\end{tabular}} \\ \hline
 &  & IC Acc & Slot F1 & IC Acc & Slot F1 & \multicolumn{1}{c|}{IC Acc} & \multicolumn{1}{c|}{Slot F1} & \multicolumn{1}{c|}{IC Acc} & \multicolumn{1}{c|}{Slot F1} \\ \hline
\citet{krone2020learning} & - & 0.877 ± 0.01 & 0.597 ± 0.017 & 0.660 ± 0.02  & 0.340 ± 0.004  & 0.877 ± 0.01  & 0.621 ± 0.007  & 0.719 ± 0.01  & 0.412 ± 0.02  \\ \hline
\emph{Baseline} (Ours) & - & 0.887 ± 0.06  &  0.597 ± 0.04 &  0.737 ±  0.06 & 0.74 ±  0.01
  & 0.907 ± 0.05
 &0.593 ± 0.04  & 0.80 ±  0.04
  &  0.70±  0.02
\\ \hline
CL (IC) & Support(m-train) &  0.905 ± 0.05 & 0.594 ± 0.04 & 0.75 ± 0.07 & 0.748 ±  0.02
  & 0.912 ±  0.03 & 0.594 ±  0.04 & 0.802 ±  0.06
 &  0.70 ±  0.02
 \\ \hline
CL (IC) & Support,Query(m-train) & 0.908 ± 0.06
 & 0.596 ± 0.04
 & \textbf{0.76 ±  0.04}
 & 0.748 ±  0.02
 & 0.93 ± 0.05
  & 0.60 ± 0.03
 & 0.829 ±  0.06
 &  0.703 ±  0.03
\\ \hline
CL (IC + SF) & Support(m-train) & 0.903 ±  0.06 & 0.60± 0.04 & 0.757 ±  0.04
 & 0.755 ±  0.02 & 0.92 ± 0.01  & 0.60 ± 0.04 &0.826 ±  0.05
  &0.70 ±  0.03
  \\ \hline
CL (IC + SF) & Support,Query(m-train) & \textbf{0.91 ± 0.04}
 & \textbf{0.60 ± 0.03}
  & 0.75 ±  0.07
 &  \textbf{0.756 ±  0.02}
& \textbf{0.93 ± 0.03}
 & \textbf{0.60 ± 0.04}
 & \textbf{0.833 ±  0.05}
  & \textbf{0.71 ±  0.02}
 \\ \hline
CL (IC + SF), DA (Slot list) & Support,Query(m-train)  & \textbf{0.921± 0.037}  & \textbf{0.619± 0.037}
& \textbf{0.803 ± 0.069} & 0.748 ± 0.019
& 0.923± 0.055  & \textbf{0.619± 0.035}
& 0.821± 0.08 & \textbf{0.73± 0.02} 
 \\ \hline
\end{tabular}}
\caption{\label{cl_table} Few-shot classification accuracy with contrastive learning (CL) for prototypical networks. For CL (IC) only $L_{contrastiveIC}$ is used, whereas for CL (IC + SF), both $L_{contrastiveIC}$ and $L_{contrastiveSF}$ are used.}
\vspace{-0.5em}
\end{table*}

We follow the few-shot learning setup for IC/SF described in \citep{krone2020learning} with a few modifications. Instead
of using a frozen backbone such as BERT or ELMo with a BiLSTM head, we use a more powerful pre-trained RoBERTa encoder.
Additionally, in contrast to \citep{krone2020learning}, we update our encoder during the meta-training stage. For a given utterance
$x^{i} = \{x^{i}_{1}, x^{i}_{2},...,x^{i}_{n} \} $ with $n$ tokens, we first use the RoBERTa model denoted by $f_{\phi}$
to encode the utterance resulting in $h^{i} = \{h^{i}_{<cls>}, h^{i}_{1},...,h^{i}_{n}\}$. We use the \texttt{<cls>}
token embedding to denote the utterance level embedding which we use for intent classification. For slot filling, we use
each of the token embeddings $\{h_{j}^{i}\}_{j=1}^{n}$ of the $i^{th}$ utterance. Given a support set $S$, assuming
$S_{l}$ consists of utterances belonging to the intent class $c_{l}$ and $S_{a}$ consists of tokens from the slot class
$c_{a}$, we first compute the class prototypes for intents ($c_{l}$) and slots ($c_{a})$:

 \begin{equation}
 \label{eqn:intent_proto}
     c_{l} = \frac{1}{|S_{l}|} \sum_{x^{i} \in S_{l}}^{}f_{\phi}(x^{i}) 
 \end{equation}
 \begin{equation}
 \label{eqn:slot_proto}
      c_{a} = \frac{1}{|S_{a}|} \sum_{x^{i}_{j} \in S_{a}}^{}f_{\phi}(x^{i}_{j}) \quad
      \forall x^{i} \in S 
 \end{equation}

Given a query example $\textbf{z}$ and a distance function $d$, a distribution over the different classes is computed
using the softmax of the distances to the different class prototypes. Specifically we denote the intent specific log
likelihood loss as:
\begin{equation}
%p(c_{l} / \textbf{z}, \phi) 
L_{IC}(\phi, \textbf{z})= - \log\{\frac{\exp(-d(f_{\phi}(\textbf{z}), c_{l}))}{\sum_{l^{'}}^{} \exp(-d(f_{\phi}(\textbf{z}), c_{l^{'}}))}\}
\end{equation}

We use euclidean distance as the standard distance function. Similarly, we define the slot specific loss as
$L_{Slots}(\phi, \textbf{z})$. For a given query set $Q$, the cumulative loss for intents and slots is the log
likelihood averaged across all the query samples and is denoted by $L_{Total}(\phi)$:
\begin{equation}
\label{proto_eq}
L_{Total}(\phi) = \sum_{\textbf{z} \in Q}^{}\frac{1}{|Q|} \{ L_{IC}(\phi, \textbf{z}) + L_{Slots}(\phi, \textbf{z}) \}
\end{equation}
\vspace{-3em}
\subsection{Contrastive Learning}
The general idea of contrastive learning \cite{chen2020simple} is to pull together the representations of similar
samples while pushing apart the representations of dissimilar samples in an embedding space. In our work, we specifically
incorporate the supervised contrastive loss as an added regularizer with the prototypical loss computation in Eq.
(\ref{proto_eq}). In particular we identify places in the meta-training pipeline where the incorporation of the
contrastive loss is most beneficial for good generalization to few-shot classes. We devise two types of contrastive
losses for the IC/SF tasks: (a) contrastive loss for intents $L_{contrastiveIC}(\phi)$ where the \texttt{<cls>} token
embedding is used in the loss; (b) contrastive loss for slots $L_{contrastiveSF}(\phi)$ where the individual token
embeddings are used in the loss. The regularized prototypical loss is the following: 
\begin{multline}
\label{cl_total}
   L_{Total}(\phi) = \sum_{\textbf{z} \in Q}^{}\frac{1}{|Q|} \{ L_{IC}(\phi, \textbf{z}) + L_{Slots}(\phi, \textbf{z}) \} \\ 
    + \lambda_{1} L_{contrastiveIC}(\phi) + \lambda_{2} L_{contrastiveSF}(\phi)
\end{multline}

We provide more details about the two contrastive losses in the Appendix section.
\vspace{-0.4em}
\subsection{Data Augmentation for Few-shot IC/SF}
Prior works in computer vision \cite{liu2020task, ni2020data} have shown that data augmentation is very effective in
meta-learning. In this section, we use various data augmentation strategies to improve the meta-learning pipeline for
IC/SF tasks. Data augmentation for joint IC/SF tasks in NLU is particularly challenging as the augmentation is primarily
possible at the level of intents. For intent level data augmentation, we use state-of-the-art techniques such as
backtranslation \cite{DBLP:journals/corr/abs-1904-12848} and EDA \cite{DBLP:journals/corr/abs-1901-11196} along with
prototypical networks. We also introduce a novel data augmentation technique called \texttt{slot-list values} which
effectively leverages the structure of joint IC/SF tasks. In particular, we investigate the effectiveness of these data
augmentation techniques in the meta-learning pipeline at different levels such as: (a) support at meta-training; (b)
support $+$ query at meta-training; (c) support at meta-testing; (d) combination of those. We provide details about
these augmentation methods below. 
\vspace{-0.3em}
\subsubsection{ \texttt{Slot-List Values} Augmentation}
In IC/SF datasets, certain slot types often can take on values specified in a finite list. For example, in the SNIPS
dataset the slot type \emph{facility} can take on values from the list \emph{["smoking room", "spa", "indoor",
"outdoor", "pool", "internet", "parking", "wifi"] }. Specific to the discrete slot filling task,
\citep{slot_example_encoder} used such values to learn an additional attention module for improving SF. Such lists can
be created from the training dataset and be used for data augmentation. We leverage such lists to create synthetic
utterances by replacing the values of slot types in a given utterance with other values from the list: e.g. given an
utterance ``\emph{Book a table at a \underline{pool} bar}'', we synthesize another utterance ``\emph{Book a table at a
\underline{indoor} bar}''.
%\vspace{-0.2em}
\subsubsection{Augmentation by Backtranslation}

Backtranslation is a technique of translating an utterance into an intermediate language and back to its original
language using a neural machine translation model. Previous work \cite{backtranslate_scale, backtranslation_attention,
backtranslation_monolingual} showed that backtranslation is extremely effective as a data augmentation technique for NLP
applications. In our paper in particular, we use a pre-trained \texttt{en-es} NMT model \cite{mariannmt} for generating
the augmented utterances. To ensure that the generated utterances are diverse, we follow the procedure in
\citep{DBLP:journals/corr/abs-1904-12848} in which we employ restricted sampling from the model output probability
distribution instead of beam-search.
\subsubsection{ EDA Data Augmentation} 

Adding small perturbations to the training data via random insertion, deletion, swapping and synonym replacement is one
simple technique to generate synthetic data for data augmentation. Previous work by \citep{wei2019eda} showed that this
EDA technique achieves state-of-the-art results on various text-classification tasks. In
our work, we use EDA to generate synthetic data to perform data augmentation at different stages of meta-learning.

\begin{table*}[t]
\hspace{1.8em}
\scalebox{0.68}{
\begin{tabular}{|c|c|c|c|c|c|l|l|l|l|}
\hline
 & Level & \multicolumn{1}{c|}{SNIPS(Kmax=20)} & \multicolumn{1}{c|}{ATIS (Kmax=20)} & \multicolumn{1}{c|}{SNIPS (Kmax=100)} & \multicolumn{1}{c|}{ATIS(Kmax=100)} \\ \hline
 &  & IC Acc  & IC Acc  & \multicolumn{1}{c|}{IC Acc}  & \multicolumn{1}{c|}{IC Acc}  \\ \hline
\cite{krone2020learning} & - & 0.877 ± 0.01  & 0.660 ± 0.02  & 0.877 ± 0.01 & 0.719 ± 0.01 \\ \hline
\emph{Baseline} (Ours) & - & 0.887 ± 0.06
 & 0.737 ±  0.06  & 0.907 ± 0.05
 &0.80 ±  0.04
 \\ \hline
DA (Slot-list) & Support(m-train) & 0.898 ± 0.061   
& 0.735 ± 0.052 
& 0.916 ± 0.055
& 0.810 ± 0.052
 \\ \hline
DA (Slot-list) & Support,Query(m-train) & 0.919 ± 0.062  
& \textbf{0.800 ± 0.054}  
& 0.917 ± 0.051 
& 0.806 ± 0.066 \\ \hline
DA (Slot-list) & Support(m-train, m-test) & 0.905± 0.062  
& 0.772 ± 0.044
& 0.922± 0.051 
& 0.818± 0.056 
 \\ \hline
DA (Slot-list) & Support(m-test) & \textbf{0.926 ± 0.038}  
& 0.764 ± 0.073  
& \textbf{0.931 ± 0.037}  
& \textbf{0.840± 0.047}   \\ \hline
DA (Backtranslation) & Support(m-train) & 0.885 ± 0.03
 & 0.77 ±  0.06
  & 0.928 ± 0.029
  & 0.79 ± 0.06  \\ \hline
DA (Backtranslation) & Support(m-train, m-test) & 0.881 ± 0.03
 & 0.79 ±  0.05
 & \textbf{0.931 ± 0.030}
   & 0.795 ± 0.06  \\ \hline
DA (Backtranslation) & Support(m-test) & 0.895 ± 0.036
  & 0.71 ±  0.06
 & 0.899 ± 0.06
 & 0.77 ±  0.14
 \\ \hline
 DA (EDA) & Support(m-train) & 
 0.893 ± 0.062
  & 0.787 ± 0.07
 & 0.911 ± 0.04
 & 0.805 ± 0.08
 \\ \hline
  DA (EDA) & Support(m-train,m-test) & 
  0.893 ± 0.047
  & 0.761 ± 0.08
 & 0.915 ± 0.04
 & 0.808 ± 0.10 
 \\ \hline
  DA (EDA) & Support(m-test) & 0.892 ± 0.047
  & 0.731 ± 0.06
 & 0.915 ± 0.05
 & 0.78 ± 0.059
 \\ \hline
\end{tabular} }
\caption{\label{da_table}Few-shot IC accuracy with Data Augmentation (DA) for prototypical networks; m-train refers to meta-training and m-test refers to meta-testing}
\vspace{-1.3em}
\end{table*}
\vspace{-0.1em}
\section{Experiments}
\vspace{-0.1em}
\textbf{Datasets:} We use two well-known standard benchmarks for IC/SF tasks: SNIPS \cite{coucke2018snips} and ATIS
\cite{hemphill-etal-1990-atis}. In general, SNIPS is a more challenging dataset as it contains intents from diverse
domains. The ATIS dataset, although imbalanced, contains intents only from the \emph{Airline} domain.

\textbf{Episode Construction:} We follow the standard episode construction technique described in
\citep{krone2020learning,triantafillou2020metadataset} where the number of classes and the shots per class in each
episode are sampled dynamically. \citet{triantafillou2020metadataset} showed that this dynamic sampling
procedure helps in dealing with the intent class imbalances which is present in ATIS.

\textbf{Few-shot Splits:} For the SNIPS dataset, we use 4 intent classes for meta-training and 3 intent classes for
meta-testing. Similar to \citep{krone2020learning}, we do not form a development split for SNIPS as there are only 7
intent classes and the episode construction process requires at least 3 classes in each split. For the ATIS dataset, we
first select intent classes with more than 15 examples, then use 5 intent classes for meta-training and 7 intent classes
for meta-testing. The rest of the classes are used as a development split. In \citep{krone2020learning}, the intent
classes for each split are \textit{manually chosen}. This is not representative of realistic situations where the types of
few-shot classes can vary considerably. To address this issue, we report our experiment results averaged over 5 seeds
where in each run the intent classes for each split are randomly sampled. In each experiment run, we evaluate our
results for 100 episodes sampled from the test-split. We refer to our re-implementation of \citep{krone2020learning}
with this strategy as \emph{Baseline}.

\textbf{Main Results:} Table \ref{cl_table} shows the results of experiments adding contrastive losses as a regularizer
to our baseline. Overall, we observe that across both SNIPS and ATIS datasets, using contrastive losses as a
regularizer predominantly improves IC accuracy while marginally improving SF F1 score. In particular, we notice that
using contrastive losses as a regularizer with both the support and query during meta-training leads to the best
performances. We also find that the combination of contrastive losses and data augmentation via \texttt{slot-list
values} outperforms models trained independently with only contrastive losses or data augmentation. 

Table \ref{da_table} shows the results of adding data augmentation to the few-shot IC tasks. 
We observe the techniques in general significantly improve the performance of few-shot IC, depending on the
level in the meta-learning pipeline at which the data is augmented. More specifically, for SNIPS we notice up to
4\% and 2\% gain in IC accuracy for $Kmax=20$ and $Kmax=100$ respectively. With EDA, we find that augmentation during
meta-training and meta-testing together leads to a noteworthy gain in few-shot IC performances across both SNIPS and
ATIS. In comparison, backtranslation is effective in improving the few-shot IC performance for SNIPS, when the shots per
class is higher such as in $Kmax=100$. However for ATIS, we observe a significant gain in IC only for $Kmax=20$.
Comparatively, our novel data augmentation approach \texttt{slot-list values} approach generally shows gain in IC at all
levels and shots per class. For SF, we find that data augmentation leads to only limited improvements when compared to
IC (see Appendix C for a detailed discussion). We hypothesize that this is because data augmentation occurs primarily
for intents and does not provide any explicit signal for improving SF tasks. To further understand if additional explicit signals for slots improve the SF performances, we incorporate syntactic information into the model (see Appendix D for details). We primarily notice that the addition of syntactic knowledge only leads to marginal improvements in SF tasks. We attribute this to the low shots per slot class, an artifact of the episodic sampling procedure \citep{krone2020learning}, done per intent class in the joint IC/SF setting.
\vspace{-0.6em}
\section{Conclusion}
\vspace{-0.5em}
In this work, we systematically dissect the existing meta-learning pipeline for few-shot IC/SF tasks and identify places where contrastive learning and data augmentation can be effective. Empirically we found that contrastive losses are
effective regularizers during meta-training and outperform the current state-of-the-art few-shot joint IC/SF benchmarks
across both SNIPS and ATIS. Our novel data augmentation technique, \texttt{slot-list values}, along with other
techniques such as backtranslation and EDA improve upon a strong few-shot baseline by a significant margin for few-shot
IC. Notably a combination of contrastive losses and data augmentation with the support and query examples during meta-training leads to the best performances across both SNIPS and ATIS. These semi-supervised strategies for improving few-shot IC/SF tasks create a strong benchmark and open up
possibilities on more stronger modes of meta-specific data augmentation and contrastive learning.

% Entries for the entire Anthology, followed by custom entries
\bibliography{anthology,emnlp2021}
\bibliographystyle{acl_natbib}

\appendix
\clearpage
\newpage
\newpage 

\section{Hyperparameters}

%In our experiments with the contrastive losses, we set $\lambda_{1} = \lambda_{2}$ in Eq. (\ref{cl_total}). 
For the ATIS dataset, we use the development set to tune for $\lambda_{1}$ and $\lambda_{2}$ in Eq. (\ref{cl_total}). For the SNIPS dataset, we empirically set both $\lambda_{1}$ and $\lambda_{2}$
to be 0.06 due to the lack of a development set. In our experiments with the three data augmentation strategies, we
generate synthetic utterances to exactly double the training data size for fair comparison throughout. Across all the
experiments, we meta-train the models for 50 episodes and use a learning rate of $5e-5$.

\section{On Contrastive Learning}

In our work, we use two types of contrastive losses for IC/SF tasks: (a) contrastive loss for intents
$L_{contrastiveIC}(\phi)$ where the \texttt{<cls>} token embedding is used in the loss; (b) contrastive loss for slots
$L_{contrastiveSF}(\phi)$ where the individual token embeddings from the encoder are used in the loss. In particular, we
use the supervised contrastive loss \cite{scl} and leverage the label information present in the support or support $+$
query set during meta-training. First we define the contrastive loss for intents $L_{contrastiveIC}(\phi)$: given a set
of utterances with their corresponding intent labels $S_{intents} =\{ (x_{i}, y_{i})_{i=1}^{m} \}$, assume $P(i)$ to be
a set consisting of examples from $S_{intents}$ with same labels as the $i^{th}$ example. Formally $P(i) :\{ x_{j} :
y_{j} = y_{i} \hspace{0.5em} \forall j \in [1,m] \hspace{0.5em} \& \hspace{0.5em} j \neq i\}$. The contrastive loss for
the intents $L_{contrastiveIC}(\phi)$ is defined as the following:

\begin{align}
   \hspace{-1.7em} \sum_{i=1}^{m} \hspace{-0.6em} -\log \Big \{ \frac{1}{| P(i) |}\hspace{-0.5em} \sum_{z \in P(i)}^{} \frac{\exp(f_{\phi}(x_{i})^{T} f_{\phi}(z))/\tau}{\sum_{j=1, j\neq i}^{m} \exp(f_{\phi}(x_{i})^{T} f_{\phi}(x_{j}))/\tau} \Big\}
\end{align}
\normalsize
Here $f_{\phi}(x_{i})$ denotes the \texttt{<cls>} embedding for the $i^{th}$ utterance. In case of slots, we first obtain the individual token embeddings in each utterance $x_{i} \hspace{0.5em} \forall i \in [1,m]$. Consider the total number of tokens to be $N$ in an episode and their associated embeddings' set to be $S_{slots} =\{ (h_{j},y_{j}^{'}), \hspace{0.6em} \forall j \in N \}$, where $y_{j}^{'}$ is the slot label for the $j^{th}$ token. Similar to the intents, we define the set $Q(i): \{ h_{j}: y_{j}^{'} = y_{i}^{'} \hspace{0.5em} \forall j \in [1,N] \hspace{0.5em} \& \hspace{0.5em} j \neq i \}$. Next we define the contrastive loss for the slots $L_{slots}(\phi)$ as:
\begin{align}
    \sum_{i=1}^{N} -\log \Big \{ \frac{1}{| Q(i) |}\hspace{-0.5em} \sum_{z \in Q(i)}^{} \frac{\exp(h_{i}^{T} z)/\tau}{\sum_{j=1, j\neq i}^{N} \exp(h_{i}^{T} h_{j} )/\tau} \Big\}
\end{align}
\begin{table*}[t]
\hspace{1.8em}
\scalebox{0.68}{
\begin{tabular}{|c|c|c|c|c|c|l|l|l|l|}
\hline
 & Level & \multicolumn{1}{c|}{SNIPS(Kmax=20)} & \multicolumn{1}{c|}{ATIS (Kmax=20)} & \multicolumn{1}{c|}{SNIPS (Kmax=100)} & \multicolumn{1}{c|}{ATIS(Kmax=100)} \\ \hline
 &  & Slot F1  & Slot F1  & \multicolumn{1}{c|}{Slot F1}  & \multicolumn{1}{c|}{Slot F1}  \\ \hline

%\cite{krone2020learning} & - 
%& 0.877 ± 0.01  & 0.660 ± 0.02  & 0.877 ± 0.01 & 0.719 ± 0.01 \\ \hline
\emph{Baseline} (Ours) & - & 0.599 ± 0.04
 & 0.748 ±  0.01  & 0.593 ± 0.04
 &0.703 ±  0.02
 \\ \hline
DA (Slot-list) & Support(m-train) & 0.603 ± 0.043   
& 0.738 ± 0.020 
& 0.609 ± 0.047  
& 0.713 ± 0.025 
 \\ \hline
DA (Slot-list) & Support,Query(m-train) & \textbf{0.609 ± 0.043}  
& 0.74 ± 0.02  
& 0.609 ± 0.03 
& 0.715 ± 0.02\\ \hline
DA (Slot-list) & Support(m-train, m-test) & 0.587± 0.045  
& 0.712 ± 0.026
& 0.595 ± 0.042 
& 0.686± 0.029 
 \\ \hline
DA (Slot-list) & Support(m-test) & 0.572 ± 0.036  
& 0.697 ± 0.028  
& 0.589 ± 0.042
& 0.684± 0.02   \\ \hline
DA (Backtranslation) & Support(m-train) & 0.595 ± 0.04
 & 0.742 ±  0.01
  & \textbf{0.611 ± 0.036}
  & \textbf{0.716 ± 0.02}  \\ \hline
DA (Backtranslation) & Support(m-train, m-test) & 0.595 ± 0.04
 & 0.742 ±  0.01
 & \textbf{0.611 ± 0.03}
   & 0.716 ± 0.02  \\ \hline
DA(Backtranslation) & Support(m-test) & 0.598 ± 0.03
  & 0.74 ±  0.01
 & 0.60 ± 0.03
 & 0.72 ±  0.01
 \\ \hline
 DA(EDA) & Support(m-train) & 
 0.585 ± 0.032
  & 0.742 ± 0.02
 & 0.596 ± 0.05
 & 0.701 ± 0.03
 \\ \hline
  DA(EDA) & Support(m-train,m-test) & 
  0.593 ± 0.033
  & 0.742 ± 0.02
 & 0.594 ± 0.04
 & 0.711 ± 0.005
 \\ \hline
  DA(EDA) & Support(m-test) & 0.586 ± 0.036
  & 0.74 ± 0.01
 & 0.593 ± 0.037
 & \textbf{0.714 ± 0.02}
 \\ \hline
\end{tabular} }
\caption{\label{da_table_sf}Few-shot Slot F1 with Data Augmentation (DA) for prototypical networks; m-train refers to meta-training and m-test refers to meta-testing}
\end{table*}
\begin{table*}[t]

\scalebox{0.68}{
\begin{tabular}{|l|l|l|l|l|l|l|l|l|}
\hline
 & \multicolumn{2}{l|}{SNIPS (Kmax = 20)} & \multicolumn{2}{l|}{SNIPS(Kmax=100)} & \multicolumn{2}{l|}{ATIS(Kmax=20)} & \multicolumn{2}{l|}{ATIS(Kmax=100)} \\ \hline
 & IC Acc & Slot F1 & IC Acc & Slot F1 & IC Acc & Slot F1 & IC Acc & Slot F1 \\ \hline
\emph{Baseline} (Ours) & 0.887 ± 0.06 & 0.597 ± 0.04 & 0.907 ± 0.05 &  0.593 ± 0.04 & 0.737 ± 0.06 & 0.748 ± 0.02 & 0.801 ± 0.05 &  0.703 ± 0.02 \\ \hline
Multi-task POS loss & 0.905 ± 0.04 & \textbf{0.603 ± 0.03} & \textbf{0.929 ± 0.03} & 0.595 ± 0.03 & \textbf{0.769 ± 0.06} & 0.75 ± 0.01 & 0.807 ± 0.05 & 0.711 ± 0.02  \\ \hline
With POS-tag features & 0.896 ± 0.06 & 0.592 ± 0.04 & 0.926 ± 0.03 & 0.590 ± 0.04 &  0.745 ± 0.06  & 0.747 ± 0.01 & 0.793 ± 0.09 & 0.713 ± 0.02\\ \hline
With noun-parser features & \textbf{0.912 ± 0.05} & 0.599 ± 0.04 &0.897 ± 0.05 & \textbf{0.597 ± 0.03} &  0.764 ± 0.04 & \textbf{0.755 ± 0.02} & 0.805 ± 0.07 & \textbf{0.715 ± 0.02}\\ \hline
\end{tabular}}

\caption{\label{pos_tag_exp} Effect of adding syntactic information into the joint IC/SF model}
\end{table*}

\section{Impact of Data Augmentation for Slot Filling}

Data augmentation for joint IC/SF tasks is challenging as augmentation is only possible at the level of intents.
Although data augmentation leads to large improvements in few-shot IC performances, its impact on SF tasks is limited.
From Table \ref{da_table_sf}, across the different data augmentation methods such as backtranslation, EDA and
\texttt{slot-list values}, we observe that there is no consistent improvements in SF performances across our different
experiment settings. We hypothesize that as data augmentation does not provide any direct signal to the SF task, the
improvements are insubstantial. To address this issue and provide a more direct signal to the SF task, we incorporate
part-of-speech (POS) and noun-phrase information of the different slot values into the joint IC/SF model. In the next
section, we discuss ways to incorporate these additional syntactic information into the meta-learning pipeline.

\section{Beyond Semantic Information}

Part-of-speech (POS) and noun-parser information can provide additional syntactic information about of an utterance,
thus augmenting the semantic information from the encoded tokens. In particular, POS tags can help resolve decisions for
ambiguous tokens or words. Previous work \cite{wang2020encoding} has shown that prior information from POS tags helps in
improving IC and SF tasks in the general supervised many shot setting. In our work, we use POS tags as an additional
source of information particularly for the few-shot setting. We propose two primary ways to incorporate POS tags in the
general meta-learning setting: (a) POS tag as an additional input feature; (b) Explicitly training the model to predict
POS tags via a multi-task loss.

In addition to POS tags, we also augment information about noun-phrases as an additional input feature. Noun chunks or
phrases have the potential to provide strong signals about possible spans of different slots to the underlying model,
thus improving SF performance. For example, in the utterance ``\emph{book me a table for one at blue ribbon
barbecue}''(with intent \emph{BookRestaurant}, and slots: \emph{ party\_size\_number:"one", restaurant\_name: "blue
ribbon barbecue"}), \emph{"blue ribbon barbecue"} is identified as a noun-chunk and the span information can potentially
help with the SF task for the \emph{restaurant\_name} slot. Conversely, the POS tag for ``\emph{one}'' is \emph{NUM} and
can help classify numeric words to the numeric slot \emph{party\_size\_number}.

\label{sec:appendix}

\subsection{Feature-Based Addition}

Previous works have shown that adding POS tags as features improves IC \cite{pos_feat_zhang, pos_feat_xie} as well SF
performances \cite{pos_feat_icsf} in many-shot settings. In this work we look into incorporating syntactic features in
our meta-learning pipeline. A simple idea to incorporate POS or noun-chunk tags of an utterance is to concatenate a
vector representation of them, $p^{i}_{j}$ and $\eta^{i}_{j}$ respectively, with the token embeddings
$f_{\phi}(x^{i}_{j})$. Formally, in our meta-learning pipeline, we revise Eq. (\ref{eqn:slot_proto}) for our slot
prototype:
\begin{equation}
 \label{eqn:slot_proto_concat}
      c_{a} = \frac{1}{|S_{a}|} \sum_{x^{i}_{j} \in S_{a}}^{}f_{\phi}(x^{i}_{j})\oplus p^{i}_{j}\oplus \eta^{i}_{j} 
      \quad \forall x^{i} \in S
 \end{equation}

\subsection{Multi-task POS Loss}

Although training language models distills implicitly the structural knowledge of the underlying languages
\cite{jawahar-etal-2019-bert,syntac_pos} into the model, such knowledge can be imperfect. Explicitly training
to learn structural knowledge such as POS tags \cite{wang2020encoding}, however, can help the model to improve on downstream tasks such
as IC/SF. We treat POS tagging as a token level classification problem, similar to SF. Given a support set $S$, assume
$S_{l}$ to consist of utterances belonging to the intent class $c_{l}$, $S_{a}$ to consist of tokens from the slot class
$c_{a}$ and $S_{pos}$ to consist of POS tag tokens from the class $c_{pos}$. In addition to the intent class prototypes
$c_{l}$ and slot class prototypes $c_{a}$, we define an additional class prototype $c_{pos}$ for the POS tags:

 \begin{equation}
 \label{eqn:pos_proto}
      c_{pos} = \frac{1}{|S_{pos}|} \sum_{x^{i}_{j} \in S_{pos}}^{}f_{\phi}(x^{i}_{j}) 
      \quad \forall x^{i} \in S
 \end{equation}
 Given a query example $\textbf{z}$, we define the corresponding loss with the POS tag prototypes as: 
  \begin{equation}
    %p(c_{l} / \textbf{z}, \phi) 
    L_{pos}(\phi, \textbf{z})= - \log\{\frac{\exp(-d(f_{\phi}(\textbf{z}), c_{pos}))}{\sum_{pos^{'}}^{} \exp(-d(f_{\phi}(\textbf{z}), c_{pos^{'}}))}\}
 \end{equation}
 For the query set $Q$, the composite loss function is the following: 
 \begin{multline}
    \label{proto_eq_pos}
     L_{Total}(\phi) = \sum_{\textbf{z} \in Q}^{}\frac{1}{|Q|} \{ L_{IC}(\phi, \textbf{z}) + L_{Slots}(\phi, \textbf{z}) \\
     + \beta L_{pos}(\phi, \textbf{z})\}
 \end{multline}
 where $\beta$ is a hyperparameter. For the ATIS dataset, we select $\beta$ by using a validation set. In case of the SNIPS dataset, we empirically set $\beta$ as 0.01 due to unavailability of a development set. 

\vspace{1em}

In Table \ref{pos_tag_exp}, we observe an improvement in both IC and SF over the baseline with the addition of information from the POS tags as an auxilliary loss. However, similar to feature-based addition, we notice only a marginal and small improvement for SF.
To understand further this issue, we exmined the episodic sampling procedure used in \citep{krone2020learning}. Across
both the SNIPS and ATIS datasets, the average shots per class for intents are $\approx 5$ and $\approx 10$ for $Kmax=20$
and $Kmax=100$ respectively. However for slots, we find that the average shots per class are $\approx$ 1.3 and $\approx$
3 for $Kmax=20$ and $Kmax=100$ respectively. We conjecture that as the shots per class for slots are much lesser in comparison to that of
intents, it results in smaller improvements when compared to intents in the joint IC/SF setting.

    \section{Compute}
For all our experiments we primarily use a V100-16GB GPU. For meta-training on ATIS for $Kmax=100$ with data augmentation, we use V100-32GB GPU due to increased memory requirements.

\end{document}